\documentclass{article}
\usepackage{ijcai25}

\usepackage{times}
\usepackage{soul}
\usepackage{url}
\usepackage[hidelinks]{hyperref}
\usepackage[utf8]{inputenc}
\usepackage[small]{caption}
\usepackage{graphicx}
\usepackage{amsthm}
\usepackage{booktabs}
\usepackage{algorithm}
\usepackage{algorithmic}
\usepackage[switch]{lineno}
\usepackage{amsmath,epsfig}
\usepackage{makecell}
\usepackage{amssymb}
\usepackage{multirow}
\usepackage{newfloat}
\usepackage{listings}
\usepackage{booktabs}
\usepackage{geometry}
\usepackage{xcolor}
\usepackage{pifont}
\title{CQVPR: Landmark-aware Contextual Queries for Visual Place Recognition}

\author{
Dongyue Li$^1$
\and
Daisuke Deguchi$^1$\and
Hiroshi Murase$^{1}$
\affiliations
$^1$Nagoya University\\
}

\begin{document}

\maketitle

\begin{abstract}
    Visual Place Recognition (VPR) aims to estimate the location of the given query image within a database of geo-tagged images. To identify the exact location in an image, detecting landmarks is crucial. However, in some scenarios, such as urban environments, there are numerous landmarks, such as various modern buildings, and the landmarks in different cities often exhibit high visual similarity. Therefore, it is essential not only to leverage the landmarks but also to consider the contextual information surrounding them, such as whether there are trees, roads, or other features around the landmarks. We propose the Contextual Query VPR (CQVPR), which integrates contextual information with detailed pixel-level visual features. By leveraging a set of learnable contextual queries, our method automatically learns the high-level contexts with respect to landmarks and their surrounding areas. Heatmaps depicting regions that each query attends to serve as context-aware features, offering cues that could enhance the understanding of each scene. We further propose a query matching loss to supervise the extraction process of contextual queries. Extensive experiments on several datasets demonstrate that the proposed method outperforms other state-of-the-art methods, especially in challenging scenarios.
\end{abstract}

\section{Introduction}

Visual Place Recognition (VPR), also referred to as image localization~\cite{localization} or visual geo-localization~\cite{geo-localization}, involves estimating the approximate location of a query image by finding its closest match within a database of geo-tagged images. In recent years, VPR has attracted significant research interest, driven by its extensive practical applications across diverse domains, including robotics~\cite{robotic1,robot3,robot4}, augmented reality~\cite{AR}, and absolute pose estimation~\cite{pose}.
Despite its wide range of applications, VPR faces several significant challenges, including variations in conditions (e.g., illumination and weather), changes in viewpoint, and perceptual aliasing~\cite{alaising}, which complicates distinguishing visually similar images from different places. These challenges highlight the complexity and importance of developing robust VPR methods for real-world scenarios.

\begin{figure}[t]
  \centering
  \includegraphics[width=\linewidth]{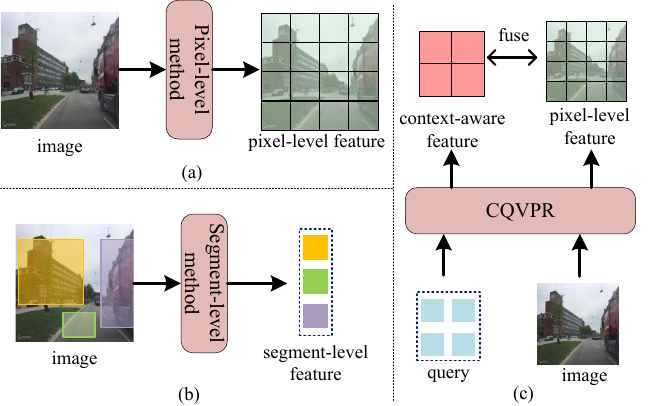}
  \caption{Conceptual difference among three VPR pipelines. (a) Pixel-level methods. (b) Segment-level methods. (c) The proposed CQVPR.}
  \label{fig0}
\end{figure}

\begin{figure}[t]
  \centering
  \includegraphics[width=0.9\linewidth]{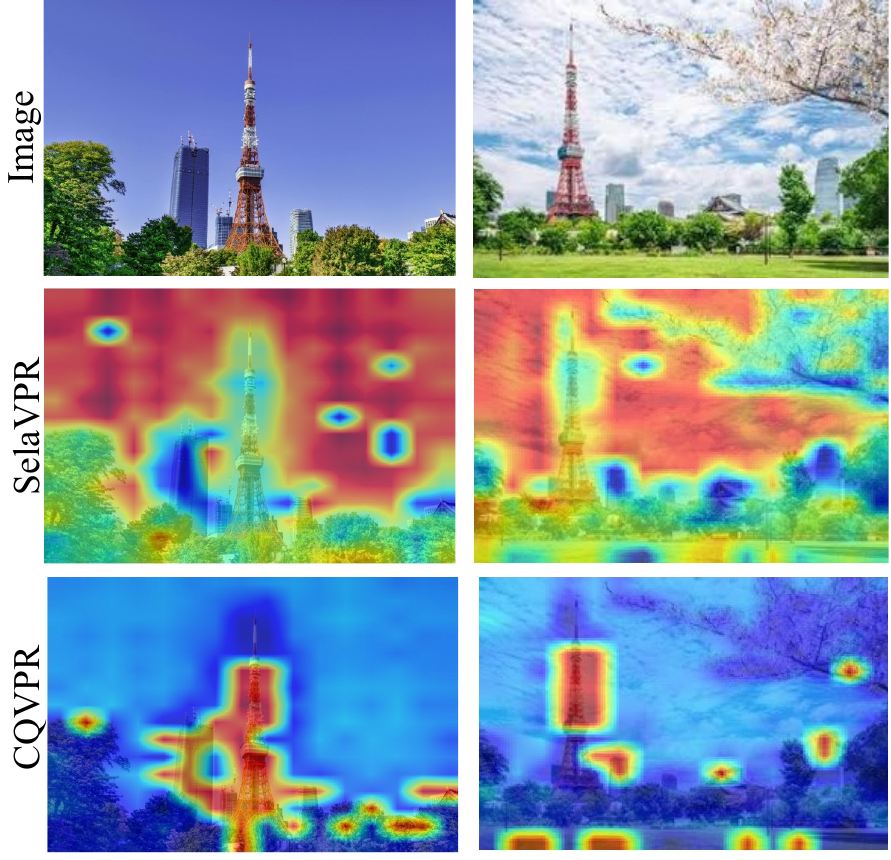}
  \caption{Comparison of cross attention maps between the previous pixel-level method SelaVPR and our CQVPR. The presented two images are from the same place. Thanks to the introduced high-level contexts, CQVPR focuses on discriminative regions (e.g., buildings and towers). While SelaVPR focuses on less informative regions.}
  \label{fig1}
\end{figure}

In VPR, the identification of key landmarks plays a pivotal role. Pixel-level methods \cite{transvpr,selavpr}, which rely exclusively on visual cues, attempt to recognize places based on image appearance. However, these approaches often lack high-level contexts, leading to an undue focus on non-landmark regions, as demonstrated in Figure \ref{fig1}. Conversely, segment-level methods often semantically partition an image into multiple clusters \cite{NetVlad} or utilize semantic segmentation models to divide an image into several objects \cite{segeccv}, which typically correspond to components of landmarks, thereby facilitating the VPR task. Nonetheless, the urban environment, which is the main environment in existing VPR datasets, poses unique challenges due to the abundance of landmarks and the high degree of visual and semantic similarity among them across different places. For instance, the Tokyo Tower and the Eiffel Tower exhibit similar appearance and semantic characteristics, despite belonging to distinct cities. Such similarities introduce significant difficulties when attempting to rely solely on segment-level or pixel-level landmark features for accurate place recognition.

Humans recognize places not only by relying on the appearance and semantics of landmarks but also by considering surrounding objects, such as the presence of nearby trees or proximity to streets. Inspired by the above fact, we propose the Contextual Query VPR (CQVPR), a method that can capture both the landmarks and the contextual information surrounding them. Figure \ref{fig0} illustrates the conceptual difference of our method and previous works.
Specifically, It leverages a set of learnable queries \cite{topicfm} to encode high-level contextual information about both the landmarks and their surroundings, where each contextual query represents a latent high-level context such as certain objects or structural shapes. Visualization results are present in Section \ref{query}. The heatmap, which depicts the regions that each query concentrates on, is fused with pixel-level features for producing global and local retrieving descriptors. Additionally, to enable the network to learn more discriminative queries, we designed a loss function that encourages the query embeddings of images from the same place to be as similar as possible, while ensuring that the query embeddings of images from different places are as dissimilar as possible. Experimental results demonstrate the proposed CQVPR can achieve accurate visual place recognition results, even in challenging scenarios involving large scale and viewpoint variations.
To summarize, the main contributions of this work are as follows:

(1) Through a learnable Transformer module, the task-related contextual queries are extracted. These inferred queries, containing rich high-level contexts, can be transformed into context-aware features for more effective visual place recognition.

(2) Contextual queries of images from the same place should be similar since they depict the same scene. Therefore, a query matching loss is proposed to maximize the similarity of queries between images from the same place while minimizing it conversely.

(3) Extensive experiments on various well-known benchmark datasets demonstrate that our proposed method outperforms several state-of-the-art baseline methods under different scenarios.

\section{Related Works}
\begin{figure*}
  \centering
  \includegraphics[width=\linewidth]{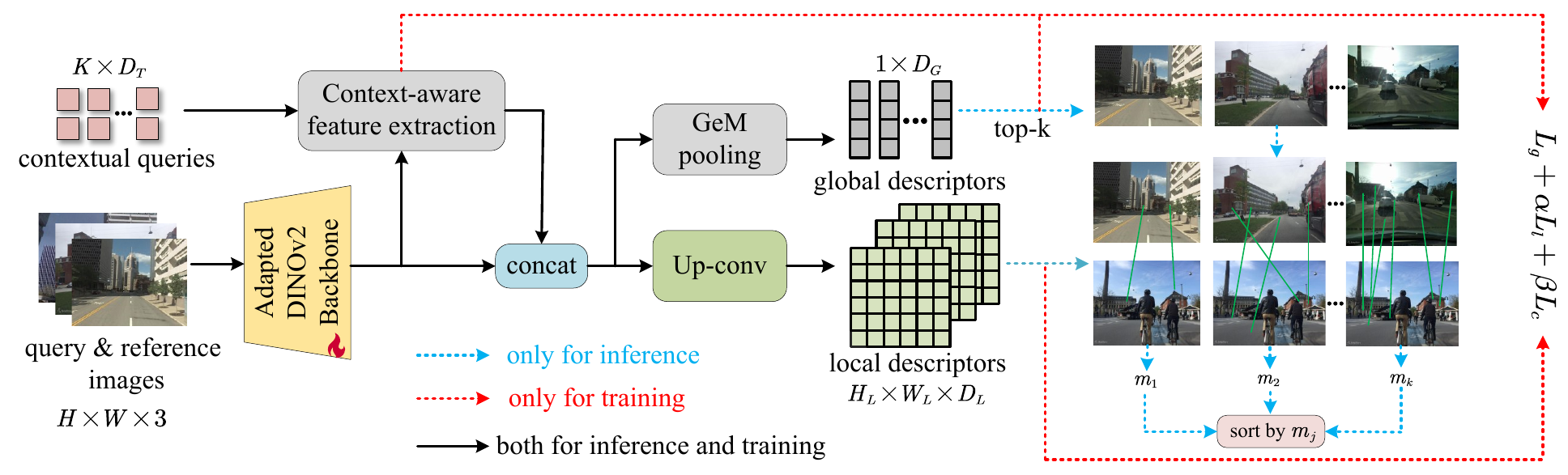}
  \caption{The overview of the proposed CQVPR. Learnable contextual queries are first randomly initialized and then transformed into context-aware features. The context-aware features are then fused with pixel-level features from the backbone.}
  \label{fig2}
\end{figure*}
\subsection{Pixel-level methods}
Pixel-level VPR methods are significant approaches in VPR. These methods typically begin by generating pixel-level features, followed by a global retrieval stage. The global retrieval stage aims to efficiently retrieve the top-k candidate images from a large database using global feature representations. The global feature is obtained by pooling pixel-level features across the entire image. For example, CricaVPR \cite{cricavpr} generates the pixel-level features through the cross-view interaction and then pools these pixel-level features into a global vector.

In addition to the global retrieval stage, some methods include a re-ranking stage. The re-ranking stage involves refining the global retrieved results by performing local feature matching between the query image and the top-k candidates. Local features are often pixel-level features or their upsampled versions. The re-ranking score between two images is determined by the number of matched points between them. The final re-ranking prediction is derived from this score. For example, AANet \cite{aanet} proposes an algorithm to align the local features under spatial constraints. $R^2$Former \cite{R2Former}
proposes a unified retrieval and re-ranking framework with only Transformers.

Recently, DINOv2 \cite{DINOv2} has achieved impressive performance in the VPR task. SelaVPR \cite{selavpr} achieves the state-of-the-art performance through fine-tuning the DINOv2's features to attend to more distinctive regions. AnyLoc \cite{anyloc} directly adopt the DINOv2 as backbone to establish an universal VPR solution. However, these methods still do not incorporate the explicit high-level contexts, making their descriptors overly rely on visual cues and can not well focus on landmarks during the local matching process.

\subsection{Segment-level methods}

Different from pixel-level methods, segment-level methods focus on generating features at a more abstract level, such as object-level, cluster-level and region-level. The global feature of an image can be seen as a special case of segment-level features, where the entire image is treated as a single segment. 

In the early stage of VPR, aggregation algorithms like VLAD~\cite{VLAD} and Bag of Words (BoW)~\cite{BoW} treat an image as a set of cluster centroids and generate cluster-level features, which are then aggregated for obtaining the final global feature representation. NetVLAD \cite{NetVlad} is proposed to make the VLAD algorithm differentiable, allowing it to be seamlessly integrated into any neural network. Patch-NetVLAD \cite{pacthnetvlad} tends to assign the NetVLAD pooled feature to each patch of the image. BoQ \cite{BoQ}, which is similar to our method, views an image as a set of queries and directly outputs the combination of these queries as the global feature. Although BoQ also introduces learned queries, it can not generate pixel-wise features and therefore can not do local matching. Specifically, the queries in BoQ are processed into global features. Due to the discrete nature of these queries, generating local features is infeasible. In contrast, our proposed CQVPR can generate both global and local features, providing a more comprehensive and versatile representation.

Recently, some methods have leveraged the explicit semantic segmentation model to generate object-level features. For example, \cite{lost} adopts the semantic segmentation backbone to generate features at each semantic class. Similar to Patch-NetVLAD, SegVLAD \cite{segeccv} leverages the recent SAM \cite{SAM} model to assign the NetVLAD pooled features to each semantic object and directly do matching at the object level. 
Despite the impressive performance achieved by these segment-level methods, their features lack spatial, appearance, and contextual information, which are crucial for distinguishing between different landmarks.


\section{Methodology}
Figure \ref{fig2} provides a detailed overview of the Contextual Query VPR (CQVPR) pipeline. Humans recognize places not just by landmarks' appearance and semantics, but also by their surrounding context, like nearby objects, trees, and streets. Building on this perspective, CQVPR is proposed to bridge the gap between pixel-level and segment-level approaches by integrating visual features with contextual information through a novel mechanism. CQVPR achieves this by leveraging a set of learnable queries to encode high-level latent contexts within an image. Each query captures specific contextual features, such as object shapes or structural elements, providing a broader understanding of the scene beyond just the landmarks. 

\subsection{Adapted DINOv2 Backbone}

DINOv2 is a pre-trained Vision Transformer (ViT)-based foundational model designed for high-quality feature representation. It has demonstrated strong robustness and impressive performance in the VPR task. Although the DINOv2’s authors reported that fine-tuning the model only brings dim improvements, the previous VPR work \cite{selavpr} found that the fine-tuned DINOv2 can bring the additional performance improvement. Therefore, we fine-tune the DINOv2 on the VPR-related datasets through adding lightweight adapters \cite{adaptformer}. The fine-tuned DINOv2 is adopted as the backbone to extract the pixel-level feature $\textbf{F}_\textbf{V}$. Below, we detailly describe the pixel-level feature extraction process through the fine-tuned DINOv2.

Given an input image $\textbf{I} \in \mathbb{R}^{H \times W \times 3} $, the backbone first divides it into $p \times p$ patches and each patch is linearly projected into an embedding with $D_C$ dimensions, resulting in a sequence of patch tokens ${\textbf{x}_1, \textbf{x}_2, \dots, \textbf{x}_N}$, where $\textbf{x}_i \in \mathbb{R}^{D_C}, N = \frac{HW}{p^2}$. Additionally, a learnable class token $\textbf{x}_{\text{class}}$ is prepended to the patch token sequence, forming the output $\textbf{X}$  
\begin{equation}
\textbf{X} = [\textbf{x}_{\text{class}}; \textbf{x}_1, \textbf{x}_2, \dots, \textbf{x}_N] \in \mathbb{R}^{(N+1) \times D_C}.
\end{equation}
Positional encodings are added to $\textbf{X}$ to preserve spatial information, and then the position-aware token sequence is passed to a stack of Transformer blocks for updating. Assuming that the input is $\textbf{X}_{l-1}$, the specific process for the $l$-th Transformer block is formulated as follows:
\begin{equation}
\begin{aligned}
&\textbf{X}'_l = \text{MHA}(\text{LN}(\textbf{X}_{l-1})) + \textbf{X}_{l-1},\\
&\textbf{X}_l = \text{AdaptMLP}(\text{LN}(\textbf{X}'_l)) + \textbf{X}'_l,\\
\end{aligned}
\end{equation}
where \text{MHA} ($\cdot$) stands for the multi-head attention, \text{LN} ($\cdot$) is the layer normalization, $\textbf{X}_l$ is the output of the $l$-th Transformer block. \text{AdaptMLP} ($\cdot$) denotes the lightweight adapter \cite{adaptformer} for fine-tuning. In the training process, only the adapters are trainable while others are frozen.

After being updated by the Transformer blocks, the class token is emitted, and the $N$ patch tokens are selected as the pixel-level feature $\textbf{F}_\textbf{V}$, which is then combined with context-aware features for obtaining the global and local descriptors.
\begin{figure*}
  \centering
  \includegraphics[width=\linewidth]{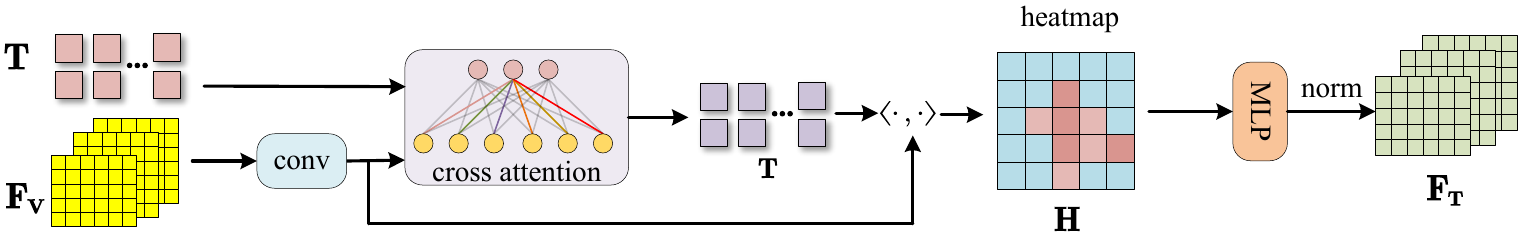}
  \caption{The illustration of the context-aware feature extraction module.}
  \label{fig3}
\end{figure*}
\subsection{Context-aware feature extraction}

We hypothesize that each image can be viewed as $K$ distinct contextual queries.
The learnable contextual embedding $\textbf{T} \in \mathbb{R}^{K \times D_T}$ representing $K$ contextual queries is first randomly initialized, and then updated through a cross attention layer,
\begin{equation}
\label{cross-att}
\textbf{T} = CA(\textbf{T}, \text{conv}(\textbf{F}_\textbf{V})),
\end{equation}
where $CA(\cdot)$ is the cross attention layer between queries $\textbf{T}$, keys $\textbf{F}_\textbf{V}$, and values $\textbf{F}_\textbf{V}$. $\textbf{F}_\textbf{V}$ is the pixel-level feature mentioned above and conv($\cdot$) stands for the convolution layer. Since $D_C$, the number of channels of $\textbf{F}_\textbf{V}$, is very large, the convolution layer here is employed to reduce it to $D_T$ for efficiency. The heatmap $\textbf{H}$ is generated through computing the similarity between $\textbf{F}_\textbf{V}$ and $\textbf{T}$.
\begin{equation}
\textbf{H} = \langle \textbf{T}, \text{conv}(\textbf{F}_\textbf{V}) \rangle,
\end{equation}
where $\langle\cdot,\cdot\rangle$ denotes the dot product and conv($\cdot$) stands for the same convolution layer in Eq \ref{cross-att}.
The heatmap $\textbf{H}$ is then transformed to be the context-aware feature $\textbf{F}_\textbf{T}$.

\begin{equation}
\textbf{F}_\textbf{T} = \text{MLP}(\text{norm}(\textbf{H})),
\end{equation}
where norm($\cdot$) means the normalization along the channel dimension and MLP($\cdot$) denotes the multi-layer perceptron layer.

\subsection{Global and local descriptors}
After the context-aware feature $\textbf{ F}_\textbf{T}$ and pixel-level feature $\textbf{F}_\textbf{V}$
are obtained, the global descriptor $\textbf{G}$ can be generated as follows
\begin{equation}
\textbf{G} = \text{L2}(\text{GeM}([\textbf{F}_\textbf{V}, \textbf{F}_\textbf{T}])),
\end{equation}
where L2($\cdot$) denotes the L2 normalization, and GeM($\cdot$) represents the GeM pooling \cite{gem}. [$\cdot$, $\cdot$] indicates the concatenation along the channel dimension. Based on global descriptors, a similarity search is performed in the global feature space across reference images, retrieving the top-k most similar candidate images to the query image. To obtain the final predictions, the local descriptors are leveraged for re-ranking these candidates. For each image, the local descriptor $\textbf{L}$ can be obtained through up-sampling the fused $\textbf{F}_\textbf{V}$ and $\textbf{F}_\textbf{T}$
\begin{equation}
\textbf{L} = \text{L2}(\text{up-conv}([\textbf{F}_\textbf{V}, \textbf{F}_\textbf{T}])),
\end{equation}
where the up-conv($\cdot$) is the up-convolution layer and the L2($\cdot$) denotes the L2 normalization. Local descriptors are leveraged to do local feature matching \cite{LoFTR} and the number of matches is treated to be the score for re-ranking the top-k candidates.

\subsection{Loss functions}
To optimize the model for generating global descriptors, a global loss \(L_g\) based on the triplet loss \cite{NetVlad} is proposed to weakly supervise the overall network,

\begin{equation}
\label{global_loss}
L_g = \sum_j l(\|\textbf{G}_q - \textbf{G}_p\| + m - \|\textbf{G}_q - \textbf{G}_{n,j}\|),
\end{equation}
where $l(x) = \max(x, 0)$ is the hinge loss, $m$ is the margin. $\textbf{G}_q$, $\textbf{G}_p$, and $\textbf{G}_{n,j}$ represent the global descriptors of the query, positive, and negative images, respectively.

For local matching, a mutual matching loss $L_l$~\cite{selavpr} is leveraged for optimizing the network to produce local descriptors that are easier to be matched. Additionally, to better supervise the extraction of contextual queries, a contextual query matching loss $L_c$ is introduced. When two images are from the same location, the similarity between their learned contextual embeddings is enlarged, whereas for images from different locations, the similarity is reduced. 
\begin{equation}
\begin{aligned}
&L_c = \sum_k l(s_{n,k} - s_p),\\
& s_p = \frac{1}{|\textbf{M}_t|}\sum_{(i, j) \in \textbf{M}_t} \textbf{T}^{T}_q(i)\textbf{T}_p(j), \\
&s_{n,k} = \frac{1}{|\textbf{M}_t'|}\sum_{(i', j') \in \textbf{M}_t'}\textbf{T}^{T}_q(i')\textbf{T}_{n,k}(j'),\\
&\textbf{M}_t = \{\left(i,j\right) \mid \forall\left(i,j\right) \in \operatorname{MNN}\left(\textbf{T}^{T}_q\textbf{T}_p\right)\},\\
&\textbf{M}_t' = \{\left(i',j'\right) \mid \forall\left(i',j'\right) \in \operatorname{MNN}\left(\textbf{T}^{T}_q\textbf{T}_{n,k}\right)\},
\end{aligned}
\end{equation}
where $\textbf{T}_q$, $\textbf{T}_p$, and $\textbf{T}_{n,k}$ represent the learned contextual embeddings of the query, positive, and negative images, respectively. MNN($\cdot$) denotes the mutual nearest neighbor criteria \cite{LoFTR}. $l(x)$ is the hinge loss. Finally, the overall loss $L$ can be obtained as
\begin{equation}
\label{loss}
L = L_g + \alpha L_l + \beta L_c,
\end{equation}
where $\alpha, \beta$ are the hyperparameters used to weight $L_l$ and $L_c$.

\begin{table*}
\centering
\caption{Comparison on Pitts30k, Tokyo24/7 and MSLS-val datasets. The bests results are highlighted in bold. The reported performance of BoQ is directly extracted from its original paper.}
\resizebox{\textwidth}{!}{
\begin{tabular}{lccccccccccc}
\toprule
\textbf{Method} & \multicolumn{3}{c}{\textbf{Pitts30k}} & \multicolumn{3}{c}{\textbf{Tokyo24/7}} & \multicolumn{3}{c}{\textbf{MSLS-val}} \\
\cmidrule(lr){2-4} \cmidrule(lr){5-7} \cmidrule(lr){8-10}
& R@1 & R@5 & R@10 & R@1 & R@5 & R@10 & R@1 & R@5 & R@10 \\
\midrule
NetVLAD\cite{NetVlad} & 81.9 & 91.2 & 93.7 & 60.6 & 68.9 & 74.6 & 53.1 & 66.5 & 71.1 \\
SFRS\cite{SFRS} & 89.4 & 94.7 & 95.9 & 81.0 & 88.3 & 92.4 & 69.2 & 80.3 & 83.1 \\
Patch-NetVLAD\cite{pacthnetvlad} & 88.7 & 94.5 & 95.9 & 86.0 & 88.6 & 90.5 & 79.5 & 86.2 & 87.7 \\
CosPlace\cite{CosPlace} & 88.4 & 94.5 & 95.7 & 81.9 & 90.2 & 92.7 & 82.8 & 89.7 & 92.0 \\
TransVPR\cite{transvpr} & 89.0 & 94.9 & 96.2 & 79.0 & 82.2 & 85.1 & 86.8 & 91.2 & 92.4 \\
StructVPR\cite{StructVPR} & 90.3 & 96.0 & 97.3 & -& - &-&88.4 &94.3& 95.0 \\
GCL\cite{GCL} & 80.7 & 91.5 & 93.9 & 69.5 & 81.0 & 85.1 & 79.5 & 88.1 & 90.1 \\
MixVPR\cite{mixvpr} & 91.5 & 95.5 & 96.3 & 85.1 & 91.7 & 94.3 & 88.0 & 92.7 & 94.6 \\
EigenPlaces\cite{EigenPlaces}& 92.5 & 96.8 & 97.6 & 93.0 & 96.2 & 97.5 & 89.1 & 93.8 & 95.0 \\
$R^2$Former\cite{R2Former}& 91.1& 95.2& 96.3&88.6 &91.4& 91.7&89.7 &95.0& 96.2\\
BoQ\cite{BoQ} & 92.4 & 96.5 & 97.4 & - & - & - & 91.2 & 95.3 & 96.1 \\
SelaVPR\cite{selavpr} & 92.8 & 96.8 & 97.7 & \textbf{94.0} & \textbf{96.8} & 97.5 & 90.8 & \textbf{96.4} & \textbf{97.2} \\
CQVPR (Ours) & \textbf{93.3} & \textbf{96.9} & \textbf{98.1} & \textbf{94.0} & \textbf{96.8} & \textbf{98.1} & \textbf{91.5} & \textbf{96.4} & 97.0 \\
\bottomrule
\end{tabular}}
\label{table1}
\end{table*}

\section{Experiments}
\subsection{Datasets and Metric}
We evaluate the proposed CQVPR on several benchmark datasets that are widely used in the VPR task, including Tokyo 24/7, MSLS, Pitts30k, Pitts250k, SPED, AmsterTime and SVOX. These datasets are selected to cover diverse environments and challenging conditions such as illumination changes, viewpoint and seasonal variations.

\textbf{Tokyo 24/7} \cite{Tokyo24/7} contains approximately 76k database images and 315 query images captured in urban environments with drastic illumination changes.
\textbf{Mapillary Street-Level Sequences (MSLS)} \cite{MSLS} comprises over 1.6 million images collected across urban, suburban, and natural scenes over seven years. Following the setup in \cite{benchmark}, CQVPR is evaluated on the val set of MSLS for comparison. 
\textbf{Pittsburgh (Pitts30k) \cite{Pitts30k}}, which is the subset of Pitts250k, includes 30k reference images and 24k query images with significant viewpoint changes. The extended \textbf{Pitts250k} dataset has 8,280 queries and a gallery of 83,952 images. \textbf{SPED} \cite{SPED} consists of images collected from 1136 webcameras around the world. \textbf{AmsterTime} \cite{Amstertime} presents a distinctive challenge by using historical grayscale images as query inputs and contemporary color images as references, capturing temporal variations spanning several decades. \textbf{SVOX} distinguishes itself with its extensive coverage of diverse weather conditions, including overcast, rainy, snowy, and sunny scenarios, providing a robust benchmark for testing adaptability to meteorological variations. In this work, CQVPR is evaluated on 2 most challenging subsets: SVOX-NIGHT and SVOX-SUN.

In the experiments, the performance is measured by using Recall@N (R@N), which indicates the percentage of queries for which at least one of the N retrieved database images falls within a specified distance threshold of the ground truth. Following previous literature, a threshold of distance is usually set to 25 meters, except for AmsterTime, where the distance threshold is set to be 10 meters.

\subsection{Implementation details}
The fine-tuned DINOv2, with only a few additional trainable lightweight adapters compared to the original, is selected as the backbone for extracting pixel-level features. Given a 224×224 input image, the backbone would generate a 1024-dimensional feature $\textbf{F}_\textbf{V}$, which has the spatial resolution of $14\times14$ pixels. The number of queries, $K$, is set to 10, and the channel dimension of the contextual embeddings, $D_T$, is set to 256.
Before calculating the heatmap $\textbf{H}$, a $1\times1$ convolution is leveraged to map $\textbf{F}_\textbf{V}$ into a 256-dimensional feature for efficiency. After $\textbf{H}$ is obtained, the MLP layer would expand the channel dimension of $\textbf{H}$ to $D_T$, namely, 256. The 256-dimensional context-aware feature $\textbf{F}_\textbf{T}$ and 1024-dimensional pixel-level feature $\textbf{F}_\textbf{V}$ are concatenated along the channel dimension to establish the 1280-dimensional feature.

To obtain the global descriptor, the GeM pooling \cite{gem} is leveraged, which is a general pooling mechanism. For the local descriptor, two $3\times3$ up-convolutions with a stride of 2 and padding of 1 are employed to up-sample the 1280-dimensional feature, resulting in a 128-dimensional local descriptor with a spatial resolution of $61\times61$ pixels. In the re-ranking process, the top-100 candidates are re-ranked to obtain final results.

CQVPR is trained using the Adam optimizer, configured with a learning rate of $10^{-5}$ and a batch size of 4. Training is terminated when the Recall@5 (R@5) on the validation set fails to improve for three consecutive epochs.
The training procedure defines positive images as reference images located within 10 meters of the query image, while definite negatives are those positioned beyond 25 meters. The margin parameter $m$ in Eq \ref{global_loss} is set to 0.1, and $\alpha$, $\beta$ in Eq \ref{loss} are both set to 1. 
\begin{table}[t]
\centering
\caption{Comparison on Pitts250k and SPED datasets. The bests results are highlighted in bold. The reported BoQ's performance is directly extracted from its original paper.}
\resizebox{\columnwidth}{!}{
\begin{tabular}{lcccc}
\toprule
\textbf{Method} & \multicolumn{2}{c}{\textbf{Pitts250k}} & \multicolumn{2}{c}{\textbf{SPED}} \\
\cmidrule(lr){2-3} \cmidrule(lr){4-5}
 & R@1 & R@5 & R@1 & R@5 \\
\midrule
NetVLAD\cite{NetVlad} & 90.5 & 96.2 & 78.7 & 88.3 \\
GeM\cite{gem} & 87.0 & 94.4 & 66.7 & 83.4 \\
Conv-AP\cite{convap}& 92.9 & 97.7 & 79.2 & 88.6 \\
CosPlace\cite{CosPlace} & 92.1 & 97.5 & 80.1 & 89.6 \\
MixVPR\cite{mixvpr} & 94.6 & 98.3 & 85.2 & 92.1 \\
EigenPlaces\cite{EigenPlaces} & 94.1 & 98.0 & 69.9 & 82.9 \\
BoQ\cite{BoQ} & 95.0 & 98.5 & 86.5& 93.4 \\
SelaVPR\cite{selavpr} & 95.7 & \textbf{98.8} & 89.0 & 94.6 \\
CQVPR (Ours) & \textbf{96.0} & 98.7 & \textbf{89.1} & \textbf{95.1} \\
\bottomrule
\end{tabular}
}
\label{table2}
\end{table}

\begin{table}[t]
\centering
\caption{Comparison (R@1) on more challenging datasets.}
\resizebox{\columnwidth}{!}{
\begin{tabular}{lccc}
\toprule
\textbf{Method} & \textbf{AmsterTime} & \textbf{SVOX-NIGHT} & \textbf{SVOX-SUN} \\
\midrule
SFRS\cite{SFRS} & 29.7 & 28.6 & 54.8 \\
CosPlace\cite{CosPlace} & 38.7 & 44.8 & 67.3 \\
MixVPR\cite{mixvpr} & 40.2 & 64.4 & 84.8 \\
EigenPlaces\cite{EigenPlaces} & 48.9 & 58.9 & 86.4 \\
SelaVPR\cite{selavpr} & 54.6 & 88.8 &  90.9\\
CQVPR (Ours) & \textbf{55.8} & \textbf{90.3} & \textbf{94.1}\\
\bottomrule
\end{tabular}
}
\label{table3}
\end{table}
CQVPR is first trained on MSLS and then subsequently trained on Pitts30k. For evaluation on MSLS-val, performance of the model only trained on MSLS is reported. For other datasets, the performance of the fully trained model is reported.

\subsection{Comparison with State-of-the-Art Methods}
\label{comparison}



As shown in Table \ref{table1}, CQVPR achieves the highest R@1 of 93.3 on Pitts30k, surpassing all other methods, including the recent SOTA method SelaVPR. On MSLS-val, which includes some suburban or natural scene images and is therefore prone to perceptual aliasing, our CQVPR still can achieve the best R@1, demonstrating its ability to produce more discriminative global and local feature descriptors to differentiate similar images from different places.
CQVPR also excels in the Pitts250k and SPED datasets, as shown in Table \ref{table2}. It achieves the highest R@1 on Pitts250k, which demonstrates its ability of being employed in the large-scale datasets.

In Table \ref{table1} and \ref{table2}, besides the improvement of the CQVPR, it is also worth highlighting the metrics saturation observed in all the above five datasets. 
Therefore, the proposed CQVPR is additionally evaluated on more challenging datasets AmsterTime, SVOX-NIGHT and SVOX-SUN. The proposed CQVPR achieves improvements of +1.2\%, +1.5\%, and +3.2\% in R@1 across these three datasets compared to the second-best method, demonstrating its superior performance in challenging scenarios with severe modality changes and illumination variations. This phenomenon proves that the introduced high-level contexts can improve the robustness of feature descriptors. 
\begin{figure}[t]
  \centering
  \includegraphics[width=\linewidth]{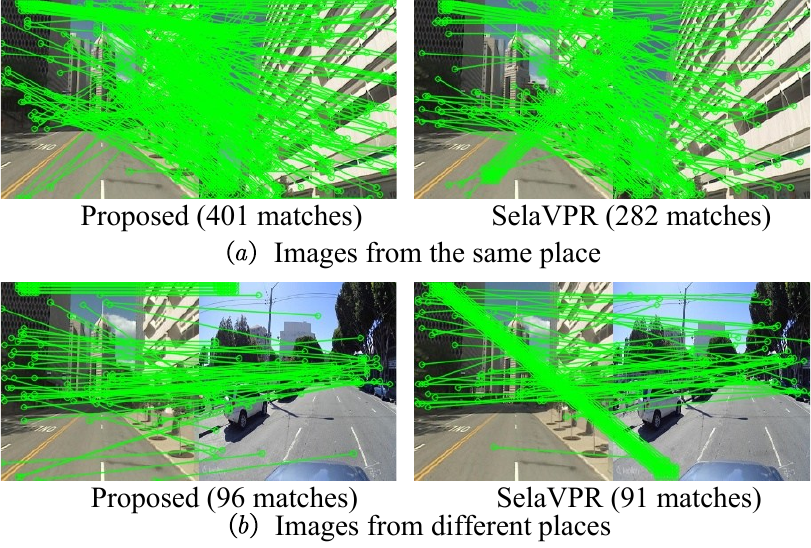}
  \caption{Comparison of local matching between CQVPR and SelaVPR.  (a) presents the local matching between images from the same place. The more matches means the better performance. (b) presents local matching between images from different places. The fewer matches the better.}
  \label{fig4}
\end{figure}

\subsection{Visualization results}
\subsubsection{Qualitative results of local matching}
In this section, we present the qualitative local matching results of our CQVPR method compared to SelaVPR, as shown in Figure \ref{fig4}. This section focuses exclusively on evaluating local matching during the re-ranking process. Therefore, homography verification is not employed here to ensure a fair comparison, as neither CQVPR nor SelaVPR uses it during the re-ranking process for efficiency.

For two images from the same place, the more matches, the better. Conversely, for two images from different places, fewer matches are preferred since all matches in this case are incorrect. As shown in Figure \ref{fig4}, our method extracts a significantly higher number of matches (1.5 times that of SelaVPR) between images from the same place. Meanwhile, for images from different places, CQVPR produces a similar number of matches as SelaVPR.

\subsubsection{Visualization of the heatmaps}
\label{query}
\begin{figure}[t]
  \centering
  \includegraphics[width=\linewidth]{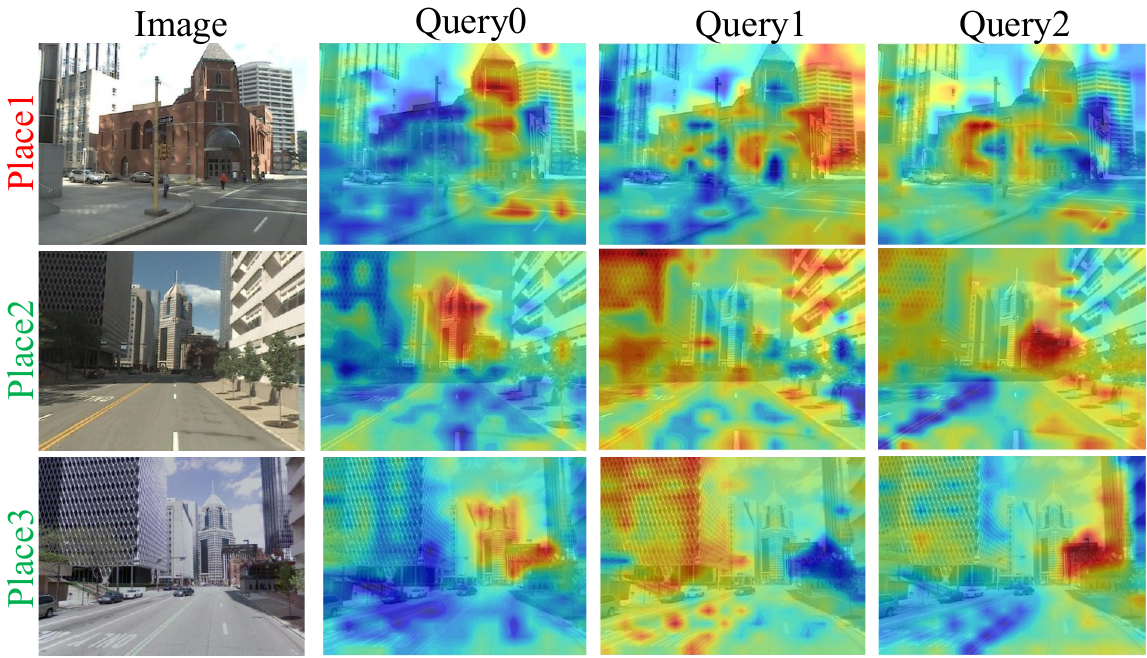}
  \caption{Visualization of the heatmap. For clarity, the number of queries is set to 3. The place 2 and place 3 are from the same place while place 1 and place 2 are from different places.}
  \label{fig6}
\end{figure}
Figure \ref{fig6} illustrates the visualization of the heatmap $\textbf{H}$, highlighting the latent semantic regions that each query individually focuses on. It can be found that each query not only corresponds to the landmark alone, but takes the surrounding environment into account. This is different from the segment-level methods. Queries of images from the same place focus on similar regions, as both describe the same scene. Conversely, heatmaps of images from different places show no overlap, as these images correspond to entirely different places.

\begin{figure*}
  \centering
  \includegraphics[width=0.9\linewidth]{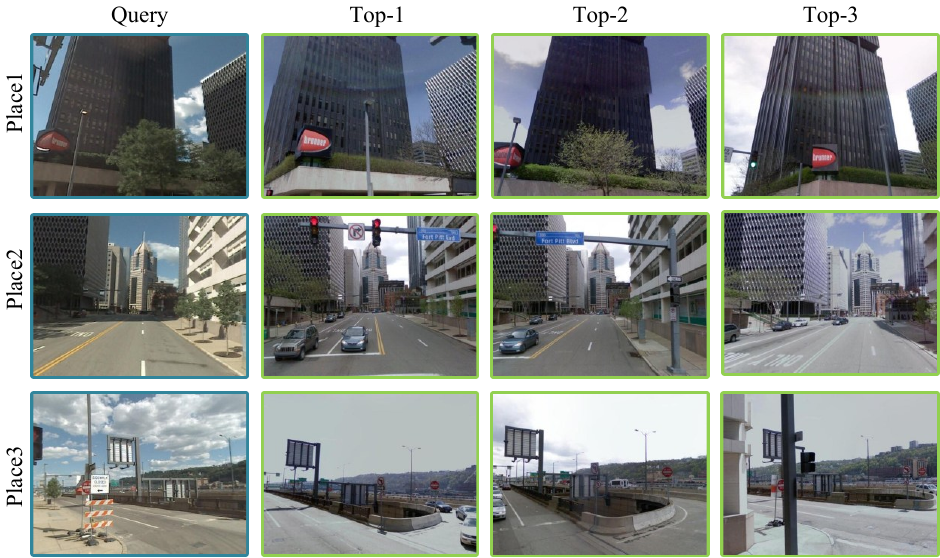}
  \caption{Qualitative results of CQVPR on the Pitts30k dataset are presented. The leftmost column displays several query images, while the subsequent three columns show the top-3 retrieved candidates for each query. All the predicted images are from the same place as the query image.}
  \label{fig5}
\end{figure*}

\begin{table}[t]
\centering
\setlength{\tabcolsep}{3pt}
\caption{Ablation on the effectiveness of each contribution.}
\resizebox{\columnwidth}{!}{
\begin{tabular}{cccccccc}
    \hline
    \multirow{2}{*}{DINOv2} & 
    \multirow{2}{*}{\makecell{fine-tuned \\ DINOv2}} & 
    \multirow{2}{*}{\makecell{context-aware \\ features}} & 
    \multirow{2}{*}{\makecell{query match- \\ ing loss}} & 
    \multicolumn{4}{c}{Pitts30k} \\
                             & & & & R@1 & R@5 & R@10 & R@20 \\
    \hline
      \ding{51}& &  &  & 87.8 & 93.8 & 96.3 & 97.6 \\
      &\ding{51} &  &  & 92.8 & 96.8 & 97.7 & 98.4 \\
      & & \ding{51} & \ding{51} & 81.2 & 92.0 & 93.9 & 95.3 \\
      &\ding{51} & \ding{51} &  & 93.0 & 96.8 & 97.8 & 98.7 \\
      & \ding{51} & \ding{51} & \ding{51} & \textbf{93.3} & \textbf{96.9} & \textbf{98.1} & \textbf{98.9} \\
    \hline
\end{tabular}
}
\label{table4}
\end{table}

\subsection{Abaltion Study}
In this section, a series of ablation experiments are conducted to better understand CQVPR. Experiments are conducted under the same training and evaluation protocol as in section \ref{comparison}.
\subsubsection{Ablation on the effectiveness of each contribution}
In this section, an ablation study is conducted on Pitts30k to verify the effectiveness of each of our contributions, as shown in Table~\ref{table4}. The third row of Table \ref{table4} presents results of only using context-aware features to do VPR. The single pixel-level or context-aware feature can not achieve the optimal performance. These results demonstrate the complementary nature of our contributions.

\begin{table}[t]
\centering
\setlength{\tabcolsep}{3pt}
\caption{Ablation on the choice of context-aware features.}
\setlength{\tabcolsep}{7pt}
\resizebox{0.9\columnwidth}{!}{
\begin{tabular}{cccccc}
    \hline
     \multirow{2}{*}{\makecell{context-aware \\ feature}} & \multicolumn{4}{c}{Pitts30k} \\
                              & R@1 & R@5 & R@10 & R@20 \\
                         \hline
    
    $\textbf{F}_\textbf{T}^*$ & 92.0 & 96.7 &97.7 &98.6 \\
    $\textbf{F}_\textbf{T}$ & \textbf{93.3} & \textbf{96.9} & \textbf{98.1} & \textbf{98.9}\\
    \hline
\end{tabular}
}
\label{table5}
\end{table}
\subsubsection{Ablation on the choice of context-aware features}
Besides processing the heatmap $\textbf{H}$ to be the context-aware feature $\textbf{F}_\textbf{T}$, in this section, we try another way to obtain the context-aware feature
\begin{equation}
\textbf{F}_\textbf{T}^* = \text{softmax}(\textbf{H})\cdot\textbf{T},
\end{equation}
where $\textbf{F}_\textbf{T}^*$ directly aggregates each query's embedding with a weighted summation. As shown in Table \ref{table5}, employing $\mathbf{F}_\mathbf{T}^*$ results in worse performance. We attribute this to the fact that directly aggregating the embedding of each query through the weighted summation may reduce the robustness of the features.

\subsubsection{Ablation on the number of queries}
In this section, we analyze the effect of the number of queries. As shown in Table \ref{table6}, setting the number of queries to 10 achieves the best performance. The reason for this phenomenon may be that the number of queries should be consistent with the number of landmarks. Since existing VPR datasets are mainly about urban scenes, an excessive or insufficient number of queries could degrade the performance.


\begin{table}[t]
\centering
\setlength{\tabcolsep}{3pt}
\caption{Ablation on the number of queries.}
\setlength{\tabcolsep}{7pt}
\resizebox{0.9\columnwidth}{!}{
\begin{tabular}{ccccc}
    \hline
     \multirow{2}{*}{\makecell{The number \\of queries}}& \multicolumn{4}{c}{Pitts30k} \\
                             & R@1 & R@5 & R@10 & R@20 \\
                         \hline
    
    5 & 92.3 & 96.6 & 97.8& 98.8\\
    10 & \textbf{93.3} & \textbf{96.9} & \textbf{98.1} & \textbf{98.9}\\
    20 & 92.6 & 96.5& 97.8& 98.6\\
    \hline
\end{tabular}
}
\label{table6}
\end{table}

\begin{table}[t]
\centering
\setlength{\tabcolsep}{3pt}
\caption{Efficiency analysis of each component for feature extraction in CQVPR.}
\setlength{\tabcolsep}{7pt}
\resizebox{0.9\columnwidth}{!}{
\begin{tabular}{ccc}
    \hline
     Module & Params (M) & Runtime (ms) \\
                         \hline
    
    fine-tuned DINOv2 & 354.77 & 27.0\\
    context-aware feature &  0.656 & 0.9 \\
    up-convolution & 3.24 & 1.0\\
    \hline
\end{tabular}
}
\label{table7}
\end{table}

\subsubsection{Efficiency analysis}
In this section, we analyze the efficiency of each component for feature extraction in CQVPR. Since the local matching process is only leveraged during inference and follows a standard procedure, it is not included in the comparison. Notably, the GeM pooling operation is excluded from the analysis as its parameters and runtime are negligible. The results in Table \ref{table7} demonstrate that the fine-tuned DINOv2 module constitutes the majority of the computational cost. Notably, when referred to Table \ref{table4}, the context-aware feature extraction module not only achieves high accuracy independently but also is highly efficient compared to the fine-tuned DINOv2.

\section{Conclusion}
In this work, we propose the Contextual Query VPR (CQVPR), which integrates contextual information with detailed pixel-level features. By introducing learnable contextual queries, our method effectively captures high-level contextual information about landmarks and their surrounding environments. Furthermore, we propose a query matching loss to supervise the context extraction process, ensuring robust and accurate context modeling.
Extensive experiments conducted on multiple datasets demonstrate CQVPR's superior performance compared to SOTA methods.
\bibliographystyle{named}
\bibliography{ijcai25}

\end{document}